%% file: main.tex
\documentclass[10pt,twocolumn,letterpaper]{article}

\usepackage[pagenumbers]{cvpr} 

\usepackage{indentfirst}
\usepackage{multirow}
\usepackage[normalem]{ulem}
\useunder{\uline}{\ul}{}
\usepackage{graphicx}
\usepackage[accsupp]{axessibility} 

\input{preamble}

\definecolor{cvprblue}{rgb}{0.21,0.49,0.74}
\usepackage[pagebackref,breaklinks,colorlinks,citecolor=cvprblue]{hyperref}

\newcommand{\boldparagraph}[1]{\noindent\textbf{#1}\ }

\def\paperID{7323} 
\def\confName{CVPR}
\def\confYear{2024}

\title{MonoCD: Monocular 3D Object Detection with Complementary Depths}

\author{
    Longfei Yan$^1$ \qquad
    Pei Yan$^1$ \qquad
    Shengzhou Xiong$^1$ \qquad
    Xuanyu Xiang$^1$ \qquad 
    Yihua Tan$^1$\thanks{Corresponding author.} \\
    $^1$Hubei Engineering Research Center of Machine Vision and Intelligent Systems, \\School of Artificial Intelligence and Automation, Huazhong University of Science and Technology, China \\
    {\tt\small \{longfeiyan, yanpei\}@hust.edu.cn, xiongshengzhou@126.com, \{xuanyuxiang, yhtan\}@hust.edu.cn}
}

\begin{document}
\maketitle
\input{sec/0_abstract}    
\input{ylf_sec/1_intro}
\input{ylf_sec/2_related}
\input{ylf_sec/3_appro}
\input{ylf_sec/4_exp}
\input{ylf_sec/5_conclu}
{
    \small
    \bibliographystyle{ieeenat_fullname}
    \bibliography{main}
}

\input{ylf_sec/6_suppl}

\end{document}

%% file: preamble.tex
\usepackage[dvipsnames]{xcolor}


%% file: sec/0_abstract.tex
\begin{abstract}
Monocular 3D object detection has attracted widespread attention due to its potential to accurately obtain object 3D localization from a single image at a low cost. Depth estimation is an essential but challenging subtask of monocular 3D object detection due to the ill-posedness of 2D to 3D mapping. Many methods explore multiple local depth clues such as object heights and keypoints and then formulate the object depth estimation as an ensemble of multiple depth predictions to mitigate the insufficiency of single-depth information. However, the errors of existing multiple depths tend to have the same sign, which hinders them from neutralizing each other and limits the overall accuracy of combined depth. To alleviate this problem, we propose to increase \textbf{the complementarity} of depths with two novel designs. First, we add a new depth prediction branch named complementary depth that utilizes global and efficient depth clues from the entire image rather than the local clues to reduce the similarity of depth predictions. Second, we propose to fully exploit the geometric relations between multiple depth clues to achieve complementarity in form. Benefiting from these designs, our method achieves higher complementarity. Experiments on the KITTI benchmark demonstrate that our method achieves state-of-the-art performance without introducing extra data. In addition, complementary depth can also be a lightweight and plug-and-play module to boost multiple existing monocular 3d object detectors. Code is available at \url{https://github.com/elvintanhust/MonoCD}.
\end{abstract}

%% file: ylf_sec/1_intro.tex
\vspace{-0.4cm}
\section{Introduction}
\label{sec:introduction}

\begin{figure}[t]
   \centering
   \includegraphics[width=0.9\linewidth]{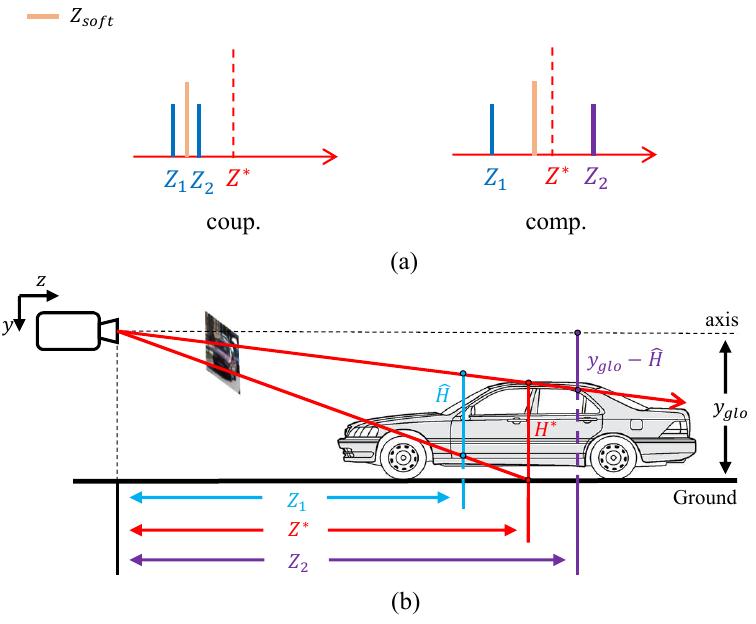}

   \caption{(a) Comparison of coupling(coup) and complementary(comp) multi-depth with two depth branches $Z_1$ and $Z_2$, where $Z^*$ and $Z_{soft}$ represents the ground truth of the depth and the final combined depth respectively. (b) A complementary demonstration of the two depth branches with the help of geometrical relations when considering only the inaccurate estimation of the object 3D height $H$. Both \textcolor[RGB]{0,176,240}{$Z_1$} generated by the widely used local height clue and \textcolor[RGB]{112,48,160}{$Z_2$} generated by our newly introduced global clue $y_{glo}$ are related to $H$. $H^*$ and $\hat{H}$ denote the ground truth of $H$ and the underestimated $H$ respectively.}
   \label{fig:1}
   \vspace{-15pt}
\end{figure}

As a significant research topic in both academia and industry, 3D object detection can empower non-human intelligences to perceive the 3D world. Compared with LiDAR-based \cite{PointPillar,PV-RCNN,PV-RCNN++,BtcDet} and stereo-based \cite{rts3d,stereo_centernet,side,Stereo_R-CNN} approaches, monocular 3D object detection has attracted widespread attention due to its lower price and simpler configuration~\cite{monodde,survey}. However, its 3D localization accuracy is significantly lower than those based on LiDAR and stereo. To advance and promote automation technologies, such as autonomous driving and robotics, it is essential to enhance the 3D localization precision of monocular 3D object detection.

Recently, many monocular 3D object detection algorithms have realized that the main reason limiting the 3D localization precision of monocular 3D object detection is inaccurate depth estimation~\cite{monoground,monoedge,DID-M3D,monoflex,monodde}. Following mainstream CenterNet paradigm~\cite{centernet}, they explore multiple local depth clues and formulate depth estimation as an ensemble of multiple depth predictions to mitigate the insufficiency of single-depth information. For instance, MonoFlex~\cite{monoflex} explores local depth clues from direct estimate and object heights, and subsequently combines them into one depth by weighted averaging. MonoDDE~\cite{monodde} further reveals clues from the object perspective point on top of that.

However, experiments on KITTI dataset~\cite{kitti} show that 95\% of the existing multi-depth prediction ensembles have the same error sign, \ie, multiple predicted depths are usually distributed on the same side of the ground truth as shown by the coupling in \cref{fig:1}(a), which leads to depth errors that cannot be neutralized with each other, hindering the improvement of combined depth accuracy. We attribute this coupling phenomenon to the fact that the local depth clues they used are all derived from the same local features around the object in the CenterNet paradigm.

In this paper, we propose to increase \textbf{the complementarity} of depths to alleviate the problem. Complementarity here refers that these predictions not only aim for high accuracy but also have different error signs. To this end, we propose two novel designs. First, considering the aforementioned coupling phenomenon, we add a new depth prediction branch that utilizes global and efficient depth clues from the entire image rather than the local clues to reduce the similarity of depth predictions. It relies on the global information that all objects in one image approximately lie on the same plane. Second, to further improve complementarity, we propose to fully exploit the geometric relations between multiple depth clues to achieve complementarity in form, which utilizes the fact that errors in the same geometric quantity may have opposite effects on different branches. For example, in \cref{fig:1}(b), $Z_1$ has a negative error because the related clue 3D height $H$ is underestimated, whereas in this case, $Z_2$ has a positive error because the effect of $H$ on $Z_2$ combined with new clues $y_{glo}$ is opposite to $Z_1$. Therefore, the geometric relation based on $H$ provides complementarity to $Z_1$ and $Z_2$ in form.

Incorporating all the designs, we propose a novel monocular 3D detector with complementary depths, named MonoCD, which compensates for the complementarity neglected in previous multi-depth predictions. The main contributions of this paper are summarized as follows:
\begin{itemize} 
   \item We point out the coupling of existing monocular object depth predictions, which limits the accuracy of the combined depths. Therefore we propose to improve the depths complementarity to alleviate this problem.
   \item We propose to add a new depth prediction branch named complementary depth that utilizes global and efficient depth clue and fully exploit the geometric relations between multiple depth clues to achieve complementarity in form.
   \item Evaluated on KITTI benchmark, our method achieves state-of-the-art performance without introducing extra data. Moreover, complementary depth can be a lightweight plug-and-play module to boost multiple existing detectors.
\end{itemize}


%% file: ylf_sec/2_related.tex
\section{Related work}
\label{sec:related}

\subsection{Center-based Monocular 3D  Detector}
\label{sec:center_based}

Many recent works \cite{centerpoint,monopair,zhou2021monoef,gupnet,monojsg,dae} are extended from the popular center-based paradigm CenterNet~\cite{centernet}, which is an anchor-free method initially applied to 2D object detection. It makes the detection process simpler and more efficient due to converting all attributes of a 3D bounding box into a center to estimate. SMOKE~\cite{smoke} inherits the center-based framework and proposes that the estimation of the 2D bounding box can be omitted. MonoDLE~\cite{monodle} finds that the estimation of the 2D bounding box contributes to the prediction of 3D attributes and demonstrates that depth error is the main reason limiting the accuracy of monocular 3D object detection. MonoCon~\cite{monocon} finds that adding auxiliary learning tasks around the center can improve the generalization performance. Although there are many benefits in the center-based framework, it makes the prediction of all 3D attributes highly correlated with the local center. It ignores the exploitation of global information, leading to the coupling of predicted 3D attributes.

\subsection{Transformer-based Monocular 3D Detector}

Benefiting from the non-local encoding of attention mechanism~\cite{attention} and its development in object detection~\cite{DETR}, multiple Transformer-based monocular 3D detectors have recently been proposed to enhance the global perception capability. MonoDTR~\cite{monodtr} proposes to perform depth position encoding to inject global depth information into Transformer to guide the detection, which requires LIDAR for auxiliary supervision. Different from it, MonoDETR~\cite{monodetr} uses foreground object labels to predict foreground Depth Maps to achieve depth guidance. In order to improve the inference efficiency, MonoATT~\cite{monoatt} proposes an adaptive token Transformer and makes it possible for finer tokens to be assigned to more significant regions in images. Although the above methods perform well, the drawbacks of high computational complexity and slow inference of Transformer-based monocular 3D detectors are still apparent. Thus there is currently a lack of a method that has both the capability of synthesizing global information and low latency in real-world autonomous driving scenarios.



\subsection{Estimation of Multi-Depth}

In addition to directly estimating object depth using deep neural networks, many recent works have broadened the depth estimation branch by mediately predicting geometric clues associated with depth. \cite{monorcnn,gupnet} utilizes mathematical priors and uncertainty modeling to restore depth information through the ratio of 3D to 2D height. Based on them, MonoFlex~\cite{monoflex} further extends the geometric depths to three sets by other supporting lines of the 3D bounding box and proposes to use uncertainties as weights to combine multiple depths into a final depth. MonoGround~\cite{monoground} introduces a local ground plane prior and enriches the depth supervision sources using randomly sampling dense points in the bottom plane of each object. MonoDDE~\cite{monodde} utilizes keypoint information to expand the number of depth prediction branches to 20, highlighting the importance of depth diversity. However, the complementarity between multiple depths is hardly explored. Errors in geometric clues (such as 2D/3D height) accumulate into the corresponding depth errors. Without effective complementarity, existing depth errors cannot be neutralized.

%% file: ylf_sec/3_appro.tex
\section{Approach}
\label{sec:approach}

\begin{figure*}
   \centering
   \includegraphics[width=0.9\linewidth]{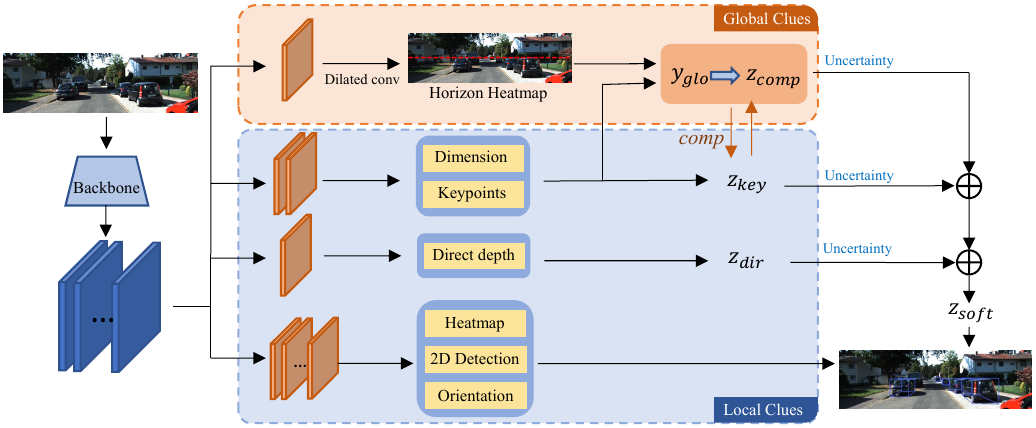}

   \caption{Overview of the approach. The input image is first subjected to processing by a feature extraction network and subsequently directed into multiple prediction heads. The prediction heads are divided into two parts. The upper orange section is used to predict the global horizon heatmap of the image, serving as a global clue to generate the prediction of complementary depths ($z_{comp}$). The lower blue section, after predicting local information for each point of interest, further generates keypoint depths ($z_{key}$) and direct depth ($z_{dir}$). Finally, the three depth prediction branches are weighted and combined using simultaneously predicted uncertainties to obtain the final depth estimation.}
   \label{fig:3}
\end{figure*}

\subsection{Problem Definition}

The task of monocular 3D object detection is to recognize objects of interest from a 2D image only and predict their corresponding 3D attributes including 3D location $(x,y,z)$, dimension $(h,w,l)$, and orientation $\theta$. The 3D location $(x,y,z)$ is usually transformed into 2.5D information $(u_c,v_c,z)$ for prediction. The recovery process of $x$ and $y$ can be formulated as:
\begin{equation}
  x=\frac{(u_c-c_u)z}{f_x},\quad y=\frac{(v_c-c_v)z}{f_y}
  \label{eq:1}
\end{equation}
where $(u_c,v_c)$ is the projected 3D center in the image and $(c_u,c_v)$ is the camera optical center. $f_x$ and $f_y$ denote the horizontal and vertical focal lengths respectively.

As described in \cref{sec:introduction}, many methods \cite{monoflex,monoground,monodde} have realized that depth $z$ is the main reason limiting the performance of monocular 3D detector and utilize multi-depth to improve the accuracy of depth prediction via: 
\begin{equation}
  z_{soft}=\displaystyle\sum_{i=1}^{n}w_iz_i
  \label{eq:2}
\end{equation}
where $\{z_i\}_{i=1}^n$ represents n predicted depths and $\{w_i\}_{i=1}^n$ represents their weights determined by the predicted uncertainty~\cite{kendall2017uncertainties,kendall2018multi}. $z_{soft}$ is used as the final depth of the output.

\subsection{The Effect of Complementary Depths}
\label{sec:prove}

To demonstrate the effectiveness of complementary depths, we present its superiority from a mathematical perspective. Define two different depth prediction branches $\hat{z}_1$ and $\hat{z}_2$ as follows:
\begin{equation}
  \hat{z}_1 = z^* + e_1,\quad \hat{z}_2 = z^* + e_2
  \label{eq:3}
\end{equation}
where $z^*$ represents the ground truth of depth. $e_1$ and $e_2$ are the errors of the two depth branches in a single prediction, respectively. Note that the positive and negative of $e_1$ and $e_2$ correspond to the sign of error. We define $e_1e_2>0$ to simulate the case of multiple depth coupling, as shown in \cref{fig:1}(a). We term the final combination error of multiple coupling depths as coupling depth error. Hence, referring to \cref{eq:2}, \textbf{the coupling depth error $E_1$} of $\hat{z}_1$ and $\hat{z}_2$ can be formulated as:
\begin{equation}
  \begin{aligned}
      E_1 &= |w_1 \hat{z}_1 + w_2 \hat{z}_2 - z^*|  \\
      &= |w_1 e_1 + w_2 e_2|
  \end{aligned}
  \label{eq:4}
\end{equation}
where $w_1$ and $w_2$ satisfy $w_1, w_2 >0$ and $w_1 + w_2 = 1$. We then flip $\hat{z}_1$ symmetrically along $z^*$ without changing the accuracy of the prediction through:
\begin{equation}
  \begin{aligned}
      \hat{z}_1^{\prime} &= z^* - (\hat{z}_1 - z^*) \\
      &= z^* - e_1
  \end{aligned}
  \label{eq:5}
\end{equation}
After flipping, the error sign in $\hat{z}_1^{\prime}$ and $\hat{z}_2$ are opposite and higher complementarity between them is artificially achieved. We term the final combination error of multiple complementary depths as complementary depth error. Similarly, \textbf{the complementary depth error $E_2$} of $\hat{z}_1^{\prime}$ and $\hat{z}_2$ can be formulated as:
\begin{equation}
  \begin{aligned}
      E_2 &= |w_1 \hat{z}_1^{\prime} + w_2 \hat{z}_2 - z^*|  \\
      &= |w_1 e_1 - w_2 e_2|
  \end{aligned}
  \label{eq:6}
\end{equation}
By mathematical transformations we further express \cref{eq:4,eq:6} as:
\begin{equation}
  \begin{aligned}
      E_1 &= \sqrt{(w_1 e_1 + w_2 e_2)^2} \\
      &= \sqrt{(w_1e_1)^2 + 2w_1w_2e_1e_2 + (w_2e_2)^2}
  \end{aligned}
  \label{eq:7}
\end{equation}

\begin{equation}
  \begin{aligned}
      E_2 &= \sqrt{(w_1 e_1 - w_2 e_2)^2} \\
      &= \sqrt{(w_1e_1)^2 - 2w_1w_2e_1e_2 + (w_2e_2)^2}
  \end{aligned}
  \label{eq:8}
\end{equation}

\begin{figure}[t]
   \centering
   \includegraphics[width=1.0\linewidth]{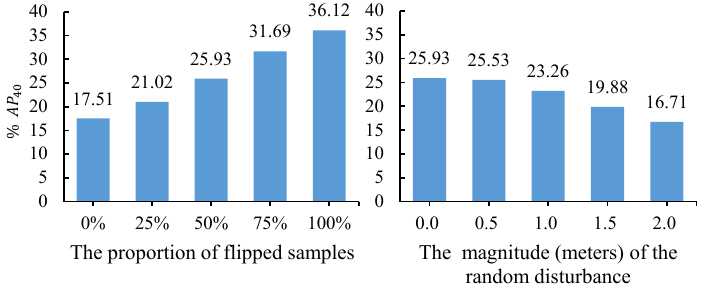}

   \caption{Evaluation of complementary effect on the KITTI \emph{validation} set. The metric is $AP_{40}$ for the moderate Car category at 0.7 IoU threshold. \emph{Left}: Different proportions of flipped samples achieve different levels of complementarity. \emph{Right}: Fixing the proportion of flipped samples to 50\% and applying random disturbances of different magnitudes to the flipped depth branch.}
   \label{fig:2}
\end{figure}

It is obvious that the complementary depth error $E_2$ is consistently less than the coupling depth error $E_1$ due to the condition $e_1e_2 > 0$.
Regardless of variations in weight or error magnitude, this relationship remains constant.
Similarly, the conclusion is equivalent by maintaining $z_1$ unchanged during the flip of $z_2$. Therefore we can draw \textbf{the conclusion}: realizing the complementary relationship between two depth branches contributes to reducing the overall depth error, even without improving the accuracy of individual branches.

To demonstrate the effectiveness of complementary depths in practice, we select a classical multi-depth prediction baseline~\cite{monoflex} for evaluation in KITTI val set. It contains 4 depth prediction branches (1 directly estimated depth and 3 geometric depths) and the coupling rate of any two branches is around 95\% after testing.
As shown on the left in \cref{fig:2}, we flip the direct depth estimation branch among them symmetrically along the ground truth based on \cref{eq:5} across a 0\% to 100\% sample scale to achieve depths complementary at different levels. Additionally, considering the difficulty of obtaining depth predictions with opposite error signs while maintaining the same accuracy in practice, we conduct another experiment by flipping the depth branch while applying random disturbances of different magnitudes on top of it. The results are presented on the right of \cref{fig:2}. Similar results are observed in other branches by performing the same operation as above. Based on this, we have the following three observations:

\textbf{Observation 1:} On the left of \cref{fig:2}, the detection accuracy increases as the proportion of flipped samples rises. It demonstrates that increasing complementarity between multiple depth prediction branches can improve detection accuracy continuously.

\textbf{Observation 2:} For two independent depth prediction branches, ideally, the proportion of their predictions with opposite signs in all samples should be 50\%. The situation is similar to the 50\% flipped proportion on the left of \cref{fig:2} due to the coupling of multiple branches in the baseline. Therefore reducing the similarity of multiple depth prediction branches can also increase their complementarity.

\textbf{Observation 3:} In the case where the flipped proportion is fixed at 50\%, as shown in the right of \cref{fig:2}, it is not until the application of random disturbance with an amplitude of 2 meters (which is quite significant~\cite{monodle} for Car in KITTI) that the complementary effect disappeared. This indicates that complementary effect can still contribute to overall performance even if losing some depth estimation accuracy and ultimately whether the overall performance can be improved depends on both the proportion of opposite signs and the depth estimation accuracy.

Additionally, we select models with different total numbers of depth prediction branches to perform flipping and evaluation. We find that as the number of flipped branches approaches the number of unflipped branches, the overall performance improves accordingly. For more experiments and details, please refer to the \textcolor{blue}{supplementary materials}.

\subsection{3D Detector with Complementary Depths}
\label{sec:detector}

\boldparagraph{Framework Overview.}
As shown in \cref{fig:3}, the network we design extends from CenterNet~\cite{centernet}. The regression heads are divided into two parts: local clues and global clues, where DLA-34~\cite{DLA} is chosen as the backbone of the network.
The branch of local clues is designed with reference to MonoFlex~\cite{monoflex}, which estimates dimension, keypoints, direct depth, orientation, and 2D detection for each local peak point based on the predicted Heatmap. Since the prediction of these geometric quantities is highly correlated with the position of the local peak point in the image, they are referred to as local clues. Both $z_{dir}$ and $z_{key}$ are derived from them. 
The branch of global clues predicts the Horizon Heatmap of the entire image based on all extracted pixel features, which is used to obtain the trend of $y_{glo}$ in scenes, and then outputs the complementary depth $z_{comp}$ embedding the global clues. How to construct a depth prediction branch with the global clues and further achieve complementarity in form will be elaborated below. Following~\cite{kendall2017uncertainties,kendall2018multi}, we model uncertainty for all seven depth predictions (1 direct depth, 3 keypoint depths, and 3 complementary depths augmented by diagonal columns as \cite{monoflex}). The final depth is obtained according to \cref{eq:2}, with $w_i=\frac{1}{\sigma_i}$.

\vspace{0.2cm}
\boldparagraph{Depth Prediction with Global Clues.}
Inspired by \cite{dijk2019neural}, the neural network sees depth from a single image through:
\begin{equation}
  z=\frac{{f}_{y}y}{v_b-c_v}
  \label{eq:10}
\end{equation}
where $y$ denotes the $y$-axis coordinates of the object in the camera coordinate system, and $v_b$ denotes the vertical coordinate of the projected bottom center in the pixel coordinate system. Considering that $y$ also represents the elevation of the plane in which the objects are located and that all objects lie approximately in one plane, $y$ contains such a global characteristic and can be distinguished from other depth clues. Unlike previous neural networks that implicitly utilize \cref{eq:10}, we propose to predict $y$ explicitly.

To avoid falling into the coupling, we do not utilize the center-based approach discussed in \cref{sec:center_based} to predict $y$. We propose to first obtain the sloping trend of $y$ in the scene by the ground plane equation. The prediction of the ground plane equation is based on the Horizon Heatmap branch, similar to \cite{gpenet}, but we omit the edge prediction and obtain prediction results as:
\begin{equation}
  \begin{aligned}
    Ax+By+Cz+1.65 = 0 \\
    s.t. \quad A^2+B^2+C^2=1
  \end{aligned}
  \label{eq:11}
\end{equation}
where $A=F\frac{k_hf_x}{f_y}$, $B=-F$ and $C=F\frac{k_hc_u+b_h-c_v}{f_y}$. $k_h$ and $b_h$ represent the slope and intercept of the horizon fitted by Horizon Heatmap. After it, then considering \cref{eq:1} and the projected bottom center $(u_b,v_b)$ of the object, $y$ with global information can be derived as:
\begin{equation}
  {y}_{glo}=-\frac{1.65}{An+Cm+B}
  \label{eq:12}
\end{equation}
where $n=\frac{f_y(u_b-c_u)}{f_x(v_b-c_v)}$, $m=\frac{f_y}{v_b-c_v}$.

Inserting \cref{eq:12} into \cref{eq:10}, a new depth prediction branch with the global clue is obtained:
\begin{equation}
  z_{glo}=\frac{{f}_{y}y_{glo}}{v_b-c_v}
  \label{eq:z_global}
\end{equation}
In addition, to better utilize the global features as well as to expand the receptive field, we use dilated convolution~\cite{yu2015multi} to predict the Horizon Heatmap.

\begin{figure}[t]
   \centering
   \includegraphics[width=1.0\linewidth]{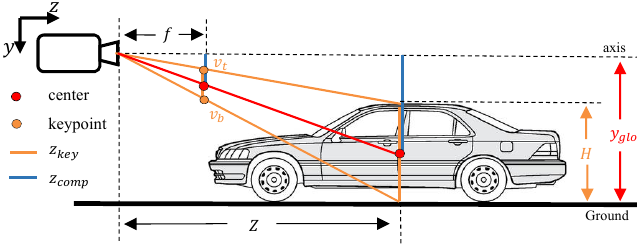}

   \caption{Geometric correspondence of different depths. To avoid overlap, the geometric correspondences of $z_{key}$ and $z_{comp}$ are marked with orange and blue lines, respectively.}
   \label{fig:4}
\end{figure}

\vspace{0.2cm}
\boldparagraph{Complementary Form in Solving.}
Simply achieving more independent depth prediction is not enough, we hope to fully exploit the geometric relations between multiple depth prediction branches to improve complementarity further. Considering the projected bottom center $(u_b,v_b)$ and top center $(u_t,v_t)$, as shown in the orange part of \cref{fig:4}, the depth derived from keypoint and height in \cite{monorcnn} can be rewritten as:
\begin{equation}
  z_{key}=\frac{f_yH}{v_b-v_t}
  \label{eq:13}
\end{equation}
where $H$ represents the 3D height of the object. Combining the global $y_{glo}$ information obtained by \cref{eq:12} and the geometric quantities used in \cref{eq:13}, we further propose a depth prediction that is complementary to $z_{key}$ in form:
\begin{equation}
  z_{comp}=\frac{f_y(y_{glo}-\frac{1}{2}H)}{\frac{1}{2}(v_b+v_t)-c_v}
  \label{eq:14}
\end{equation}
The geometric correspondence is shown in the blue part of \cref{fig:4}. It can be observed that the signs of $H$ and $v_t$ in the designed \cref{eq:14} are exactly opposite to those in \cref{eq:13}. This means that the errors of $H$ and $v_t$ have opposite effects on $z_{key}$ and $z_{comp}$ during the prediction of 3D information for each object. Although \cref{eq:13} and \cref{eq:14} are not strictly symmetrical, this further increases the probability that the errors $e_{key}$ and $e_{comp}$ of $z_{key}$ and $z_{comp}$ satisfy the condition of $e_{key}e_{comp}<0$. As proved by \cref{sec:prove}, eventually a part of the depth error is neutralized in the weighted averaging of \cref{eq:2}.

%% file: ylf_sec/4_exp.tex
\section{Experiments}

\begin{table*}
\centering
\begin{tabular}{l|c|ccc|ccc|c}
\hline
\multirow{2}{*}{Methods, Venues} & \multirow{2}{*}{Extra data} & \multicolumn{3}{c|}{Test, $AP_{3D}$} & \multicolumn{3}{c|}{Test, $AP_{BEV}$} & \multirow{2}{*}{Time(ms)} \\
 &  & Eazy & Mod. & Hard & Eazy & Mod. & Hard &  \\ \hline
DDMP-3D~\cite{ddmp-3d}, CVPR2021 & Depth & 19.71 & 12.78 & 9.80 & 28.08 & 17.89 & 13.44 & 180 \\ \hline
Kinematic3D~\cite{Kinematic3D}, ECCV2020 & Video & 19.07 & 12.72 & 9.17 & 26.69 & 17.52 & 13.10 & 120 \\ \hline
AutoShape~\cite{autoshape}, ICCV2021 & \multirow{2}{*}{CAD} & 22.47 & 14.17 & 11.36 & 30.66 & 20.08 & 15.59 & 50 \\
DCD~\cite{DCD}, ECCV2022 &  & 23.81 & 15.90 & 13.21 & 32.55 & 21.50 & 18.25 & - \\ \hline
MonoRUn~\cite{monorun}, CVPR2021 & \multirow{3}{*}{LiDAR} & 19.65 & 12.30 & 10.58 & 27.94 & 17.34 & 15.24 & 70 \\
CaDDN~\cite{caddn}, CVPR2021 &  & 19.17 & 13.41 & 11.46 & 27.94 & 18.91 & 17.19 & 630 \\
MonoDTR~\cite{monodtr}, CVPR2022 &  & 21.99 & 15.39 & 12.73 & 28.59 & 20.38 & 17.14 & 37 \\ \hline
SMOKE~\cite{smoke}, CVPRW2020 & \multirow{9}{*}{None} & 14.03 & 9.76 & 7.84 & 20.83 & 14.49 & 12.75 & 30 \\
MonoDLE~\cite{monodle}, CVPR21 &  & 17.23 & 12.26 & 10.29 & 24.79 & 18.89 & 16.00 & 40 \\
MonoRCNN~\cite{monorcnn}, ICCV2021 &  & 18.36 & 12.65 & 10.03 & 25.48 & 18.11 & 14.10 & 70 \\
MonoFlex~\cite{monoflex}, CVPR2021 &  & 19.94 & 13.89 & 12.07 & 28.23 & 19.75 & 16.89 & 35 \\
MonoGround~\cite{monoground}, CVPR2022 &  & 21.37 & 14.36 & 12.62 & 30.07 & 20.47 & 17.74 & 30 \\
GPENet~\cite{gpenet}, - &  & 22.41 & 15.44 & 12.84 & 30.31 & 20.79 & 18.21 & - \\
MonoJSG~\cite{monojsg}, CVPR2022 &  & 24.69 & 16.14 & 13.64 & 32.59 & 21.26 & 18.18 & 42 \\
MonoCon~\cite{monocon}, AAAI2022 &  & 22.50 & 16.46 & {\ul 13.95} & 31.12 & 22.10 & {\ul 19.00} & 25.8 \\
MonoDETR~\cite{monodetr}, ICCV2023 &  & {\ul 25.00} & {\ul 16.47} & 13.58 & \textbf{33.60} & {\ul 22.11} & 18.60 & 43 \\ \hline
MonoCD(Ours) & None & \textbf{25.53} & \textbf{16.59} & \textbf{14.53} & {\ul 33.41} & \textbf{22.81} & \textbf{19.57} & 36 \\
\textit{Improvement} & \textit{v.s. second-best} & +0.53 & +0.12 & +0.58 & -0.19 & +0.70 & +0.57 & - \\ \hline
\end{tabular}
\caption{Comparison with current state-of-the-art methods on Car category on the KITTI test set. Methods are grouped according to extra data. Follow~\cite{kitti}, the methods in each group are sorted by ${AP}_{3D}$ performance in Moderate difficulty setting. We \textbf{bold} the best results and {\ul underline} the second results.}
\label{tab:1}
\end{table*}

\subsection{Dataset}
Our experiments are conducted on the widely-adopted KITTI 3D Object~\cite{kitti} dataset, which contains 7481 training images and 7518 test images. Since the annotations of the test images are not publicly accessible, we follow~\cite{chen20153d} and further divide the 7481 training images into 3712 and 3769 as the training and validation sets, respectively. Each category is further refined into three difficulties: Easy, Moderate, and Hard based on 2D height, truncation, and occlusion.

\subsection{Evaluation Metrics}
As in previous methods, we use Average Precision ${AP}_{3D}$ and ${AP}_{BEV}$ as the overall evaluation metrics. Following~\cite{simonelli2019disentangling}, 40 recall positions are used for the above AP calculations. The IoU threshold is 0.7 for Car.

In the ablation study of \cref{sec:abla}, the mean absolute error (MAE) of $y$ is introduced as a metric to evaluate the accuracy of the different $y$ sources. In addition, to better measure the complementarity between different designs, we quantify the magnitude of complementarity as the Complementarity Score. As discussed in \cref{sec:prove}, both the error sign opposite proportion and depth estimation accuracy are crucial in achieving enhanced performance. Thus we formulate the \textbf{Complementarity Score}(CS) as:
\begin{equation}
  CS=\frac{ESOP_z}{{MAE}_{z}}
  \label{eq:CS}
\end{equation}
where $ESOP_z$ represents depths \textbf{Error Sign Opposite Proportion} (ESOP) between global and local clue branches, and ${MAE}_{z}$ represents the Mean Absolute Error of $z_{comp}$. For a baseline without $z_{comp}$, ESOP counts the proportion between $z_{key}$ and $z_{dir}$.

\subsection{Implementation Details}
In order to demonstrate the effectiveness of the proposed framework, we choose three recent center-based methods with excellent performance as the baseline model, MonoFlex~\cite{monoflex}, MonoDLE~\cite{monodle}, and MonoCon~\cite{monocon}. All experiments are performed on a single RTX 2080Ti GPU. The aforementioned baseline models all employ DLA-34~\cite{DLA} as the feature extraction network. In the Global Clues branch, the prediction head of Horizon Heatmap contains two 3×3 conv layers with BN and ReLU (where the dilation rate is set to 2) and an output conv layer. The horizon equation is obtained by taking out all the largest elements in each column of the Horizon heatmap and fitting them. The ground truth of Horizon Heatmap is generated by fitting the scene ground plane through the bottom coordinate annotation of each object and then projecting to the 2D image plane~\cite{gpenet}, so only RGB image data and camera annotations are used throughout the training process. The radius of the Gaussian kernel used for each pixel is 2 when mapping the horizon equation into Heatmap. The $z_{direct}$, $z_{key}$ and $z_{comp}$ loss weight proportions are set to $1:0.2:0.1$. The remaining settings such as optimizer, batch sizes, image padding size, \etc. remain consistent with the baseline.

\subsection{Quantitative Results}

To demonstrate the effectiveness of the proposed method, we conduct quantitative experiments on test and val sets of KITTI~\cite{kitti}.

As shown in \cref{tab:1}, the proposed method is compared with the state-of-the-art methods in recent years on the widely used KITTI test set. Our method achieves the best performance in the majority of metrics without using any additional data. Compared with the previous multi-depth solving method MonoFlex~\cite{monoflex}, our performance for ${AP}_{3D}/{AP}_{BEV}$ improves by 19.44\%/15.49\%, respectively. The performance for ${AP}_{3D}/{AP}_{BEV}$ improves from 15.44/20.79 to 16.59/22.81 compared to the method GPENet~\cite{gpenet}, which also incorporated the ground plane equation solution. Even when compared to the latest Transformer-based detector MonoDETR~\cite{monodetr}, we outperform it in most metrics while ensuring real-time operation.

\begin{table}
\centering
\resizebox{\columnwidth}{!}{%
\begin{tabular}{l|ccc|ccc}
\hline
 & \multicolumn{3}{c|}{Val, $AP_{BEV}$} & \multicolumn{3}{c}{Val, $AP_{3D}$} \\
\multirow{-2}{*}{Method} & Eazy & Mod. & Hard & Eazy & Mod. & Hard \\ \hline
MonoDLE~\cite{monodle} & 24.97 & 19.33 & 17.01 & 17.45 & 13.66 & 11.68 \\
\textbf{+ Ours} & \textbf{26.84} & \textbf{20.86} & \textbf{17.89} & \textbf{18.60} & \textbf{15.09} & \textbf{12.86} \\
Improvement & {\color[HTML]{0000FF} +1.87} & {\color[HTML]{0000FF} +1.53} & {\color[HTML]{0000FF} +0.88} & {\color[HTML]{0000FF} +1.15} & {\color[HTML]{0000FF} +1.43} & {\color[HTML]{0000FF} +1.18} \\ \hline
MonoFlex~\cite{monoflex} & 30.51 & 23.16 & 19.87 & 23.64 & 17.51 & 15.14 \\
\textbf{+ Ours} & \textbf{31.49} & \textbf{23.56} & \textbf{20.12} & \textbf{24.22} & \textbf{18.27} & \textbf{15.42} \\
Improvement & {\color[HTML]{0000FF} +0.98} & {\color[HTML]{0000FF} +0.40} & {\color[HTML]{0000FF} +0.25} & {\color[HTML]{0000FF} +0.58} & {\color[HTML]{0000FF} +0.76} & {\color[HTML]{0000FF} +0.28} \\ \hline
MonoCon~\cite{monocon} & 33.36 & 24.39 & 21.03 & 26.33 & 19.01 & 15.98 \\
\textbf{+ Ours} & \textbf{34.60} & \textbf{24.96} & \textbf{21.51} & \textbf{26.45} & \textbf{19.37} & \textbf{16.38} \\
Improvement & {\color[HTML]{0000FF} +1.24} & {\color[HTML]{0000FF} +0.57} & {\color[HTML]{0000FF} +0.48} & {\color[HTML]{0000FF} +0.12} & {\color[HTML]{0000FF} +0.36} & {\color[HTML]{0000FF} +0.40} \\ \hline
\end{tabular}%
}
\caption{In order to fully demonstrate the effectiveness of the proposed method, we extend complementary depth to three center-based monocular 3D detectors. Evaluation is performed on the KITTI val set. The increased performance is highlighted in \textcolor{blue}{blue}.\vspace{-2mm}}
\label{tab:2}
\end{table}

As shown in \cref{tab:2}, we extend the complementary depth branch to three competitive center-based monocular 3d detectors. The results of the KITTI val set demonstrate that the proposed complementary depth is flexible and achieves stable increments across multiple frameworks and metrics. It is worth noting that the boost of our design performs better on ${AP}_{BEV}$ than ${AP}_{3D}$ in general. 
We attribute this to the focus of our method on improvements in depth estimation, since ${AP}_{BEV}$ is more emphasis on the accuracy of localization along the Z-axis compared to ${AP}_{3D}$~\cite{kitti}.

\subsection{Ablation Study}
\label{sec:abla}

\begin{table}
\centering
\resizebox{\columnwidth}{!}{%
\begin{tabular}{l|ccc|c|c|c|c}
\hline
\multirow{2}{*}{Setting} & \multicolumn{3}{c|}{Val, $AP_{3D}$} & \multirow{2}{*}{\begin{tabular}[c]{@{}c@{}}$y$\\ MAE\end{tabular}} & \multirow{2}{*}{\begin{tabular}[c]{@{}c@{}}$z_{comp}$\\ MAE\end{tabular}} & \multirow{2}{*}{\begin{tabular}[c]{@{}c@{}}ESOP\\ (\%)\end{tabular}} & \multirow{2}{*}{{\begin{tabular}[c]{@{}c@{}}CS$\uparrow$\end{tabular}}} \\
 & Eazy & Mod. & Hard &  &  &  &  \\ \hline
Baseline & 23.64 & 17.51 & 15.14 & - & - & 4.08 & - \\ \hline
Baseline+lo. & 18.41 & 13.49 & 10.90 & 0.127 & 4.03 & 18.63 & 4.62 \\
Baseline+fi. & 21.93 & 15.86 & 13.22 & 0.250 & 8.47 & 45.72 & 5.40 \\
Baseline+gl. & 22.97 & 17.85 & 15.11 & 0.139 & 3.29 & 36.91 & 11.22 \\
Baseline+gt. & 26.21 & 19.43 & 16.50 & 0.097 & 3.23 & 59.08 & 18.29 \\ \hline
Baseline+gl.+ed. & 21.85 & 15.97 & 13.26 & 0.242 & 6.72 & 42.51 & 6.33 \\
Baseline+gl.+di. & 24.22 & 18.27 & 15.42 & 0.131 & 3.09 & 38.19 & 12.36 \\ \hline
\end{tabular}%
}
\caption{Ablation study of $y$ sources on KITTI val set. "lo." means using the local clues branch to predict $y$ for each object. "fi." means using fixed 1.65 meters as the $y$ source. "gl." means using the global clue branch to predict. "gt." means directly using the ground plane equation generated by the ground truth of val set. "ed." means using edge detection to obtain the horizon slope in the global clues branch. "di." means using dilated convolution.}
\label{tab:3}
\end{table}

In this section, we select MonoFlex~\cite{monoflex} as the baseline to discuss the impact of different designs.

\vspace{0.2cm}
\boldparagraph{Source of Depth Clue.}
To demonstrate the effectiveness of introducing global depth clue, we adopt different approaches to obtain depth clue $y$, and the results are presented in rows 2, 3, 4, and 5 of \cref{tab:3}. 
By comparing the ESOP metric, it can be observed that the ESOP of 3rd, 4th, and 5th in \cref{tab:3} with global characteristic (\ie, not determined by a single object) are significantly higher than that of the baseline and using local clue branch, which demonstrates the necessity of introducing global clues and the coupling of multi-depth prediction is alleviated. In addition, it can be found that the accuracy of $z_{comp}$ is largely related to the accuracy of $y$.

By comparing the results of $z_{comp}$ MAE and ESOP pairs under different settings, it can be found that determining whether complementary depth can lead to overall performance enhancement often requires evaluation from two perspectives: depth estimation accuracy and ESOP. This trend can be effectively quantified by complementary scores.

\begin{figure*}
   \centering
   \includegraphics[width=0.95\linewidth]{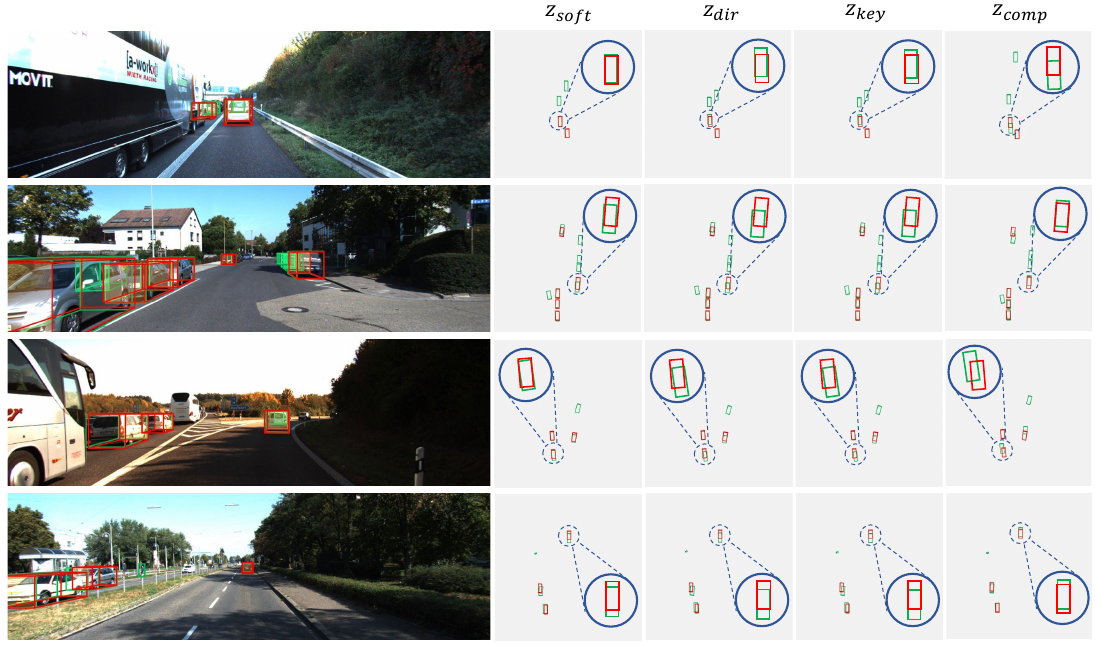}

   \caption{Qualitative examples on KITTI validation set. In each row, we provide one final front view (left) and four bird's-eye view (right) visualizations. The detection results for the various bird's-eye views vary only in terms of the depth output, progressing from $z_{soft}$ to $z_{dir}$, $z_{key}$, and $z_{comp}$ from left to right. \textcolor{red}{Red} represents the ground truth of boxes, while \textcolor[RGB]{13,173,81}{Green} represents the predictions. We circle some objects to highlight the differences across multiple depth prediction branches.}
   \label{fig:5}
   \vspace{-10pt}
\end{figure*}

The results in the 6th to 7th rows of \cref{tab:3} justify the removal of edge detection and the use of dilated convolution when predicting the ground plane equation.

\vspace{0.2cm}
\boldparagraph{Complementary Form.}
\begin{table}
\centering
\resizebox{\columnwidth}{!}{%
\begin{tabular}{l|ccc|c|c|c}
\hline
\multirow{2}{*}{\begin{tabular}[c]{@{}l@{}}Depth \\ Form\end{tabular}} & \multicolumn{3}{c|}{Val, $AP_{3D}$} & \multirow{2}{*}{\begin{tabular}[c]{@{}c@{}}$z$\\ MAE\end{tabular}} & \multirow{2}{*}{\begin{tabular}[c]{@{}c@{}}ESOP\\ (\%)\end{tabular}} & \multirow{2}{*}{{\begin{tabular}[c]{@{}c@{}}CS$\uparrow$\end{tabular}}} \\
 & Eazy & Mod. & Hard &  &  &  \\ \hline
Baseline & 23.64 & 17.51 & 15.14 & - & 4.08 & - \\ \hline
\cref{eq:z_global} & 23.16 & 17.62 & 14.73 & 2.27 & 25.69 & 11.32 \\
\cref{eq:15} & 21.83 & 15.97 & 13.19 & 8.65 & 45.40 & 5.25 \\
\cref{eq:14} & 24.22 & 18.27 & 15.42 & 3.09 & 38.19 & 12.36 \\ \hline
\end{tabular}%
}
\caption{Ablation Study of complementary forms in KITTI val set. $z$ MAE reflects the depth estimation accuracy in each form\vspace{-2mm}}
\label{tab:4}
\end{table}
To validate the effectiveness of achieving complementary form in enhancing detection accuracy, we present the results of different depth forms in \cref{tab:4}. According to the results of the 2nd and 4th row in \cref{tab:4}, the ESOP and CS of \cref{eq:14} are further enhanced after considering the complementary form compared to \cref{eq:z_global}. Although a part of the depth estimation accuracy is sacrificed, the complementarity and overall performance are eventually improved, which is consistent with observation 3 in \cref{sec:prove}.

In addition to \cref{eq:z_global,eq:14} mentioned in \cref{sec:detector}, we also consider the following complementary form:
\begin{equation}
  z=\frac{f_y(y_{glo}-H)}{v_t-c_v}
  \label{eq:15}
\end{equation}
Although it appears that \cref{eq:15} is more symmetrical and complementary to $z_{key}$ in form, its depth estimation error is significantly higher than that of \cref{eq:14}. This is due to the fact that $v_t$ and $c_v$ in the denominator are relatively close, as well as the $y_{glo}$ and $H$ in the numerator, which causes an unstable depth estimation. This is also why \cref{eq:15} has a higher ESOP because the instability of the estimate mitigates the prediction tendency, but it does not contribute to the overall performance. It demonstrates the importance of an appropriate form of complementary depth. 

\subsection{Qualitative Results}

Based on the qualitative results shown in \cref{fig:5}, it can be observed that $z_{comp}$ from the global clue branch is significantly different from $z_{dir}$ and $z_{key}$ from the local clue branch and has the opposite error sign. After combining $z_{comp}$, the predicted box is closer to the ground truth. This visualizes the process of error neutralization.

%% file: ylf_sec/5_conclu.tex
\section{Conclusion}
In this paper, we point out the coupling phenomenon that the existing multi-depth predictions tend to have the same sign, which limits the accuracy of combined depth. We analyze how complementary depth fixes it by mathematical derivation and find that the complementarity needs to be considered both from depth estimation accuracy and error sign opposite proportion. To improve depth complementarity, we propose to add a new depth prediction branch with the global clue and achieve complementarity in form through geometric relations. Extensive experiments demonstrate the effectiveness of our method.
\textbf{Limitations.} The performance of our framework is limited by the accuracy of the vertical position of objects and the complementary effect may be lost when the ground plane is undulating. Future work could involve improving the understanding and prediction of global road scenarios.
\section*{Acknowledgement}
This work was supported in part by the National Natural Science Foundation of China (No.62371201), by the Basic Research Surpport Plan of HUST (No.6142113-JCKY2022003), and by the China Scholarship Council for funding visiting Ph.D. student (No.202106160054).
\clearpage

%% file: ylf_sec/6_suppl.tex
\clearpage
\setcounter{page}{1}
\maketitlesupplementary
\appendix

\section{Cross-Dataset Evaluation}

To demonstrate the generalizability of our proposed method, we conduct cross-dataset evaluations on KITTI~\cite{kitti} and nuScenes~\cite{nuscenes} datasets. Following \cite{monorcnn}, our model is trained on the KITTI training set (3712 images), and evaluated on KITTI (3769 images) and nuScenes frontal (6019 images) validation sets. We also provide the results of retraining MonoCon~\cite{monocon} using the official code but unrestricted from training on distant objects ($z>65m$) as a fair comparison with others. To fit the model trained in KITTI, for the nuScenes dataset, we adjusted the image resolution to 384×672 and the ground plane equation prediction preset height to 1.562m (the ego car height in nuScenes~\cite{nuscenes}). Neither our method nor MonoCon uses normalized coordinates for the direct depth prediction branch and the images of KITTI and nuScenes have different focal lengths which the direct depth prediction relies on. Thus, following \cite{deviant}, we divide their direct predicted depth by 1.361.

The cross-dataset evaluation results are shown in \cref{tab:cross}, our method has lower prediction errors at different object depth ranges, which indicates the effectiveness of the proposed complementary depths in improving overall accuracy. In addition, our method outperforms other methods in most of the metrics on both datasets, which demonstrates the generalizability of our method.

\section{Discussion on multi-depth prediction methods}
\label{sec:discussion}

\cref{tab:1} shows some representative multi-depth prediction methods in recent years. The coupling between their multiple branches is shown in the third column of \cref{tab:1} in terms of Error Sign Opposite Proportions (ESOP). MonoFlex~\cite{monoflex} contains 4 depth prediction branches including 1 directly predicted depth and 3 depths shown in the 2nd row of \cref{tab:1}. MonoGround~\cite{monoground} and our method have 3 additional depth branches on top of them. Since the results of the public branches are similar, for MonoGround and our method, \cref{tab:1} only shows the results of unshared branches.

It can be observed that the error sign of the 3 depths from keypoint and height is similar to the error sign of the directly predicted depths. Benefiting from the wider range of dense depth supervision, the coupling phenomenon of depths from the ground added by MonoGround~\cite{monoground} is mitigated a bit, but it does not eliminate the coupling. Because its dense supervision comes from local sampled values around the bottom of the object. Although the code of MonoDDE~\cite{monodde} has not been released, a similar coupling phenomenon can be inferred based on the local information it uses. However, after our complementary design, the coupling phenomenon is significantly alleviated and the overall performance is further improved.

\begin{table}
\centering
\resizebox{\columnwidth}{!}{%
\begin{tabular}{l|l|cccc}
\hline
\multirow{2}{*}{Dataset} & \multirow{2}{*}{Method} & \multicolumn{4}{c}{Depth prediction MAE (meters)$\downarrow$} \\ \cline{3-6} 
 &  & 0-20 & 20-40 & \multicolumn{1}{c|}{40-$\infty$} & All \\ \hline
\multirow{5}{*}{KITTI} & M3D-RPN~\cite{m3d-rpn} & 0.56 & 1.33 & \multicolumn{1}{c|}{2.73} & 1.26 \\
 & MonoRCNN~\cite{monorcnn} & 0.46 & 1.27 & \multicolumn{1}{c|}{2.59} & 1.14 \\
 & GUPNet~\cite{gupnet} & 0.45 & 1.10 & \multicolumn{1}{c|}{1.85} & 0.89 \\
 & MonoCon~\cite{monocon} & 0.40 & 1.08 & \multicolumn{1}{c|}{1.78} & 0.85 \\
 & MonoCD(Ours) & \textbf{0.37} & \textbf{1.04} & \multicolumn{1}{c|}{\textbf{1.72}} & \textbf{0.83} \\ \hline
\multirow{5}{*}{nuScenes} & M3D-RPN~\cite{m3d-rpn} & 0.94 & 3.06 & \multicolumn{1}{c|}{10.36} & 2.67 \\
 & MonoRCNN~\cite{monorcnn} & 0.94 & 2.84 & \multicolumn{1}{c|}{8.65} & 2.39 \\
 & GUPNet~\cite{gupnet} & 0.82 & 1.70 & \multicolumn{1}{c|}{6.20} & 1.45 \\
 & MonoCon~\cite{monocon} & 0.78 & 1.65 & \multicolumn{1}{c|}{6.02} & 1.40 \\
 & MonoCD(Ours) & \textbf{0.73} & \textbf{1.59} & \multicolumn{1}{c|}{\textbf{5.78}} & \textbf{1.33} \\ \hline
\end{tabular}%
}
\caption{Cross-dataset evaluation on KITTI and nuScenes frontal validation with depth prediction MAE.}
\label{tab:cross}
\end{table}

\begin{table}
\centering
\resizebox{0.8\columnwidth}{!}{%
\begin{tabular}{l|c|c|c}
\hline
Model & \begin{tabular}[c]{@{}c@{}}Branch\\ dir\&\end{tabular} & \begin{tabular}[c]{@{}c@{}}ESOP\\ (\%)$\uparrow$\end{tabular} & \begin{tabular}[c]{@{}c@{}}Val, $AP_{3D}$\\ Mod.$\uparrow$\end{tabular} \\ \hline
\multirow{3}{*}{MonoFlex~\cite{monoflex}} & key0 & 4.08 & \multirow{3}{*}{17.51} \\
 & key1 & 5.22 &  \\
 & key2 & 6.19 &  \\ \hline
\multirow{3}{*}{MonoGround~\cite{monoground}} & gro0 & 18.35 & \multirow{3}{*}{18.69} \\
 & gro1 & 20.72 &  \\
 & gro2 & 14.73 &  \\ \hline
\multirow{3}{*}{\begin{tabular}[c]{@{}l@{}}MonoCD\\ (Ours)\end{tabular}} & comp0 & 38.19 & \multirow{3}{*}{19.37} \\
 & comp1 & 40.24 &  \\
 & comp2 & 40.05 &  \\ \hline
\end{tabular}%
}
\caption{Comparison between multiple depth prediction methods. The second column in the table represents the branches used to calculate ESOP with the directly(dir) predicted depth of each model. Including depths from keypoint and height (key), depths from ground (gro), and depths for complementary (comp). Different suffix numbers are used to distinguish the specific branches. The accuracy in the last column is $AP_{40}$ for the moderate Car category at 0.7 IoU threshold on KITTI.}
\label{tab:1}
\end{table}

\section{Additional Experiments on the Effect of Complementary Depths}
\label{sec:exp}

This section supplements the part of Sec. \textcolor{red}{3.2} in the main paper that is not presented in detail due to space limits. With the analyses in this section, two experimental conclusions can be obtained: 

(1) Existing multiple predicted depths suffer from a common problem of lacking complementarity.

(2) To maximize the complementary effect, it is beneficial to keep prediction branches symmetrical in number.

\subsection{Flip on Different Branch}

\begin{table}
\centering
\resizebox{0.8\columnwidth}{!}{%
\begin{tabular}{l|ccccc}
\hline
\multirow{2}{*}{\begin{tabular}[c]{@{}l@{}}Fipped\\ Branch\end{tabular}} & \multicolumn{5}{c}{Proportion of Flipped Samples} \\
 & 0\% & 25\% & 50\% & 75\% & 100\% \\ \hline
dir & 17.51 & 21.02 & 25.93 & 31.69 & 36.12 \\
key0 & 17.51 & 21.06 & 25.78 & 31.26 & 35.87 \\
key1 & 17.51 & 20.92 & 25.55 & 30.87 & 35.42 \\
key2 & 17.51 & 20.85 & 25.33 & 29.76 & 34.92 \\ \hline
\end{tabular}%
}
\caption{Perform flipping operation on different depth branches according to different sample proportions on KITTI dataset.}
\label{tab:2}
\end{table}

\begin{table}
\centering
\resizebox{0.75\columnwidth}{!}{%
\begin{tabular}{l|c|c}
\hline
Model & \begin{tabular}[c]{@{}c@{}}Numbers of\\ Flipped\\ Branches\end{tabular} & \begin{tabular}[c]{@{}c@{}}Val, $AP_{3D}$\\ Mod.$\uparrow$\end{tabular} \\ \hline
\multirow{5}{*}{MonoFlex~\cite{monoflex}} & 0 & 17.51 \\
 & 1 & 25.93 \\
 & 2 & 35.79 \\
 & 3 & 22.95 \\
 & 4 & 15.55 \\ \hline
\multirow{8}{*}{MonoGround~\cite{monoground}} & 0 & 18.69 \\
 & 1 & 20.59 \\
 & 2 & 21.79 \\
 & 3 & 24.24 \\
 & 4 & 32.34 \\
 & 5 & 32.75 \\
 & 6 & 22.60 \\
 & 7 & 17.12 \\ \hline
\end{tabular}%
}
\caption{Evaluation results of two multi-depth prediction models with different numbers of flipped branches on KITTI dataset, where the proportion of flipped samples is fixed at 50\%.}
\label{tab:3}
\end{table}

As shown in \cref{tab:2}, we perform flipping on different branches of MonoFlex~\cite{monoflex} according to different flipped sample proportions. The first row of results in the table is presented to the left of Fig. \textcolor{red}{3} in the main paper. It can be observed that the results of selecting different branches for flipping are similar, which indicates that the coupling between multiple-depth branches is relatively similar and lacking complementarity is common.

\subsection{Flip with Different Numbers of Branches}

To maximize the complementary effects, we additionally conducted an analytical study on two multi-depth prediction models with different numbers of flipped branches. The results in \cref{tab:3} show that realizing branch flips with different numbers is effective in improving performance except in the case where all branches are flipped. This is because although the accuracy of the depth prediction does not change with flipping, the depth values will be completely flipped to the other side of the ground truth. According to Eq. (\textcolor{red}{1}) in the main paper, it introduces additional error to the predicted $x$ and $y$, resulting in a decrease in the accuracy of the predicted 3D bounding box. 

Furthermore, it is worth noting that both models perform best when the number of flipped branches and the number of unflipped branches are close to the same. This indicates that for multiple depth prediction branches with complementary effects, maintaining a certain level of symmetry in number is preferable to maximize their effectiveness. This is why we follow the number of $z_{key}$ and design three symmetrical $z_{comp}$ in the main paper.

\begin{table}
\centering
\resizebox{\columnwidth}{!}{%
\begin{tabular}{l|cccccl}
\hline
\multirow{5}{*}{Setting} & \multicolumn{6}{c}{Combined Depth prediction MAE (meters) $\downarrow$} \\ \cline{2-7} 
 & \multicolumn{5}{c|}{$y_{glo}$ MAE (meters) $\downarrow$} & \multirow{4}{*}{overall} \\ \cline{2-6}
 & 0-0.1 & 0.1-0.2 & 0.2-0.3 & 0.3-0.4 & \multicolumn{1}{c|}{0.4-$\infty$} &  \\ \cline{2-6}
 & \multicolumn{5}{c|}{Proportion of samples (\%)} &  \\ \cline{2-6}
 & 54.09 & 27.37 & 9.61 & 4.77 & \multicolumn{1}{c|}{4.15} &  \\ \hline
Baseline & 0.90 & 1.17 & 1.72 & 1.84 & \multicolumn{1}{c|}{\textbf{2.78}} & \multicolumn{1}{c}{1.18} \\
MonoCD(Ours) & \textbf{0.85} & \textbf{1.13} & \textbf{1.66} & \textbf{1.82} & \multicolumn{1}{c|}{3.02} & \multicolumn{1}{c}{\textbf{1.14}} \\ \hline
\end{tabular}%
}
\caption{The system robustness evaluation in KITTI val set, which contains five levels based on the MAE of $y_{glo}$. The larger the value, the worse the conditions the system faces. The percentage under each level represents the proportion of samples.}
\label{tab:r1}
\end{table}

\section{System Robustness Evaluation}
As we discussed in the limitations of the main paper, the performance of our method is affected by the estimation of the ground plane equation and keypoints. Thus, we conduct a system robustness evaluation to check the performance of our method in severe conditions as shown in \cref{tab:r1}. For our added complementary depths, the effect of inaccuracies in ground plane estimation or keypoint detection is directly reflected in the prediction error of $y_{glo}$. Therefore, we divide the samples into five levels according to the MAE of $y_{glo}$ and count the mean absolute error of the combined depth at each level. It can be observed that our method outperforms the baseline in most cases, and in a few severe conditions (less than 5\%), the performance of our method degrades. This problem will be alleviated by enhancing the understanding of road scenes in the future.

%% file: main.bbl
\begin{thebibliography}{48}
\providecommand{\natexlab}[1]{#1}
\providecommand{\url}[1]{\texttt{#1}}
\expandafter\ifx\csname urlstyle\endcsname\relax
  \providecommand{\doi}[1]{doi: #1}\else
  \providecommand{\doi}{doi: \begingroup \urlstyle{rm}\Url}\fi

\bibitem[Brazil and Liu(2019)]{m3d-rpn}
Garrick Brazil and Xiaoming Liu.
\newblock M3d-rpn: Monocular 3d region proposal network for object detection.
\newblock In \emph{ICCV}, pages 9287--9296, 2019.

\bibitem[Brazil et~al.(2020)Brazil, Pons-Moll, Liu, and Schiele]{Kinematic3D}
Garrick Brazil, Gerard Pons-Moll, Xiaoming Liu, and Bernt Schiele.
\newblock Kinematic 3d object detection in monocular video.
\newblock In \emph{ECCV}, pages 135--152. Springer, 2020.

\bibitem[Caesar et~al.(2020)Caesar, Bankiti, Lang, Vora, Liong, Xu, Krishnan, Pan, Baldan, and Beijbom]{nuscenes}
Holger Caesar, Varun Bankiti, Alex~H Lang, Sourabh Vora, Venice~Erin Liong, Qiang Xu, Anush Krishnan, Yu Pan, Giancarlo Baldan, and Oscar Beijbom.
\newblock nuscenes: A multimodal dataset for autonomous driving.
\newblock In \emph{CVPR}, pages 11621--11631, 2020.

\bibitem[Carion et~al.(2020)Carion, Massa, Synnaeve, Usunier, Kirillov, and Zagoruyko]{DETR}
Nicolas Carion, Francisco Massa, Gabriel Synnaeve, Nicolas Usunier, Alexander Kirillov, and Sergey Zagoruyko.
\newblock End-to-end object detection with transformers.
\newblock In \emph{ECCV}, pages 213--229. Springer, 2020.

\bibitem[Chen et~al.(2021)Chen, Huang, Tian, Gao, and Xiong]{monorun}
Hansheng Chen, Yuyao Huang, Wei Tian, Zhong Gao, and Lu Xiong.
\newblock Monorun: Monocular 3d object detection by reconstruction and uncertainty propagation.
\newblock In \emph{CVPR}, pages 10379--10388, 2021.

\bibitem[Chen et~al.(2015)Chen, Kundu, Zhu, Berneshawi, Ma, Fidler, and Urtasun]{chen20153d}
Xiaozhi Chen, Kaustav Kundu, Yukun Zhu, Andrew~G Berneshawi, Huimin Ma, Sanja Fidler, and Raquel Urtasun.
\newblock 3d object proposals for accurate object class detection.
\newblock \emph{NeurIPS}, 28, 2015.

\bibitem[Chen et~al.(2020)Chen, Tai, Sun, and Li]{monopair}
Yongjian Chen, Lei Tai, Kai Sun, and Mingyang Li.
\newblock Monopair: Monocular 3d object detection using pairwise spatial relationships.
\newblock In \emph{CVPR}, pages 12093--12102, 2020.

\bibitem[Dijk and Croon(2019)]{dijk2019neural}
Tom~van Dijk and Guido~de Croon.
\newblock How do neural networks see depth in single images?
\newblock In \emph{ICCV}, pages 2183--2191, 2019.

\bibitem[Geiger et~al.(2012)Geiger, Lenz, and Urtasun]{kitti}
Andreas Geiger, Philip Lenz, and Raquel Urtasun.
\newblock Are we ready for autonomous driving? the kitti vision benchmark suite.
\newblock In \emph{CVPR}, pages 3354--3361. IEEE, 2012.

\bibitem[Huang et~al.(2022)Huang, Wu, Su, and Hsu]{monodtr}
Kuan-Chih Huang, Tsung-Han Wu, Hung-Ting Su, and Winston~H Hsu.
\newblock Monodtr: Monocular 3d object detection with depth-aware transformer.
\newblock In \emph{CVPR}, pages 4012--4021, 2022.

\bibitem[Kendall and Gal(2017)]{kendall2017uncertainties}
Alex Kendall and Yarin Gal.
\newblock What uncertainties do we need in bayesian deep learning for computer vision?
\newblock \emph{NeurIPS}, 30, 2017.

\bibitem[Kendall et~al.(2018)Kendall, Gal, and Cipolla]{kendall2018multi}
Alex Kendall, Yarin Gal, and Roberto Cipolla.
\newblock Multi-task learning using uncertainty to weigh losses for scene geometry and semantics.
\newblock In \emph{CVPR}, pages 7482--7491, 2018.

\bibitem[Kumar et~al.(2022)Kumar, Brazil, Corona, Parchami, and Liu]{deviant}
Abhinav Kumar, Garrick Brazil, Enrique Corona, Armin Parchami, and Xiaoming Liu.
\newblock Deviant: Depth equivariant network for monocular 3d object detection.
\newblock In \emph{ECCV}, pages 664--683. Springer, 2022.

\bibitem[Lang et~al.(2019)Lang, Vora, Caesar, Zhou, Yang, and Beijbom]{PointPillar}
Alex~H Lang, Sourabh Vora, Holger Caesar, Lubing Zhou, Jiong Yang, and Oscar Beijbom.
\newblock Pointpillars: Fast encoders for object detection from point clouds.
\newblock In \emph{CVPR}, pages 12697--12705, 2019.

\bibitem[Li et~al.(2019)Li, Chen, and Shen]{Stereo_R-CNN}
Peiliang Li, Xiaozhi Chen, and Shaojie Shen.
\newblock Stereo r-cnn based 3d object detection for autonomous driving.
\newblock In \emph{CVPR}, pages 7644--7652, 2019.

\bibitem[Li et~al.(2021)Li, Su, and Zhao]{rts3d}
Peixuan Li, Shun Su, and Huaici Zhao.
\newblock Rts3d: Real-time stereo 3d detection from 4d feature-consistency embedding space for autonomous driving.
\newblock In \emph{AAAI}, pages 1930--1939, 2021.

\bibitem[Li et~al.(2022{\natexlab{a}})Li, Chen, He, and Zhang]{DCD}
Yingyan Li, Yuntao Chen, Jiawei He, and Zhaoxiang Zhang.
\newblock Densely constrained depth estimator for monocular 3d object detection.
\newblock In \emph{ECCV}, pages 718--734. Springer, 2022{\natexlab{a}}.

\bibitem[Li et~al.(2022{\natexlab{b}})Li, Qu, Zhou, Liu, Wang, and Jiang]{monodde}
Zhuoling Li, Zhan Qu, Yang Zhou, Jianzhuang Liu, Haoqian Wang, and Lihui Jiang.
\newblock Diversity matters: Fully exploiting depth clues for reliable monocular 3d object detection.
\newblock In \emph{CVPR}, pages 2791--2800, 2022{\natexlab{b}}.

\bibitem[Lian et~al.(2022)Lian, Li, and Chen]{monojsg}
Qing Lian, Peiliang Li, and Xiaozhi Chen.
\newblock Monojsg: Joint semantic and geometric cost volume for monocular 3d object detection.
\newblock In \emph{CVPR}, pages 1070--1079, 2022.

\bibitem[Liu et~al.(2022)Liu, Xue, and Wu]{monocon}
Xianpeng Liu, Nan Xue, and Tianfu Wu.
\newblock Learning auxiliary monocular contexts helps monocular 3d object detection.
\newblock In \emph{AAAI}, pages 1810--1818, 2022.

\bibitem[Liu et~al.(2020)Liu, Wu, and T{\'o}th]{smoke}
Zechen Liu, Zizhang Wu, and Roland T{\'o}th.
\newblock Smoke: Single-stage monocular 3d object detection via keypoint estimation.
\newblock In \emph{CVPRW}, pages 996--997, 2020.

\bibitem[Liu et~al.(2021)Liu, Zhou, Lu, Fang, and Zhang]{autoshape}
Zongdai Liu, Dingfu Zhou, Feixiang Lu, Jin Fang, and Liangjun Zhang.
\newblock Autoshape: Real-time shape-aware monocular 3d object detection.
\newblock In \emph{ICCV}, pages 15641--15650, 2021.

\bibitem[Lu et~al.(2021)Lu, Ma, Yang, Zhang, Liu, Chu, Yan, and Ouyang]{gupnet}
Yan Lu, Xinzhu Ma, Lei Yang, Tianzhu Zhang, Yating Liu, Qi Chu, Junjie Yan, and Wanli Ouyang.
\newblock Geometry uncertainty projection network for monocular 3d object detection.
\newblock In \emph{ICCV}, pages 3111--3121, 2021.

\bibitem[Ma et~al.(2021)Ma, Zhang, Xu, Zhou, Yi, Li, and Ouyang]{monodle}
Xinzhu Ma, Yinmin Zhang, Dan Xu, Dongzhan Zhou, Shuai Yi, Haojie Li, and Wanli Ouyang.
\newblock Delving into localization errors for monocular 3d object detection.
\newblock In \emph{CVPR}, pages 4721--4730, 2021.

\bibitem[Peng et~al.(2022{\natexlab{a}})Peng, Wu, Yang, Liu, and Cai]{DID-M3D}
Liang Peng, Xiaopei Wu, Zheng Yang, Haifeng Liu, and Deng Cai.
\newblock Did-m3d: Decoupling instance depth for monocular 3d object detection.
\newblock In \emph{ECCV}, pages 71--88. Springer, 2022{\natexlab{a}}.

\bibitem[Peng et~al.(2022{\natexlab{b}})Peng, Zhu, Wang, and Ma]{side}
Xidong Peng, Xinge Zhu, Tai Wang, and Yuexin Ma.
\newblock Side: center-based stereo 3d detector with structure-aware instance depth estimation.
\newblock In \emph{Proceedings of the IEEE/CVF Winter Conference on Applications of Computer Vision}, pages 119--128, 2022{\natexlab{b}}.

\bibitem[Qian et~al.(2022)Qian, Lai, and Li]{survey}
Rui Qian, Xin Lai, and Xirong Li.
\newblock 3d object detection for autonomous driving: A survey.
\newblock \emph{Pattern Recognition}, 130:\penalty0 108796, 2022.

\bibitem[Qin and Li(2022)]{monoground}
Zequn Qin and Xi Li.
\newblock Monoground: Detecting monocular 3d objects from the ground.
\newblock In \emph{CVPR}, pages 3793--3802, 2022.

\bibitem[Reading et~al.(2021)Reading, Harakeh, Chae, and Waslander]{caddn}
Cody Reading, Ali Harakeh, Julia Chae, and Steven~L Waslander.
\newblock Categorical depth distribution network for monocular 3d object detection.
\newblock In \emph{CVPR}, pages 8555--8564, 2021.

\bibitem[Shi et~al.(2020)Shi, Guo, Jiang, Wang, Shi, Wang, and Li]{PV-RCNN}
Shaoshuai Shi, Chaoxu Guo, Li Jiang, Zhe Wang, Jianping Shi, Xiaogang Wang, and Hongsheng Li.
\newblock Pv-rcnn: Point-voxel feature set abstraction for 3d object detection.
\newblock In \emph{CVPR}, pages 10529--10538, 2020.

\bibitem[Shi et~al.(2023)Shi, Jiang, Deng, Wang, Guo, Shi, Wang, and Li]{PV-RCNN++}
Shaoshuai Shi, Li Jiang, Jiajun Deng, Zhe Wang, Chaoxu Guo, Jianping Shi, Xiaogang Wang, and Hongsheng Li.
\newblock Pv-rcnn++: Point-voxel feature set abstraction with local vector representation for 3d object detection.
\newblock \emph{International Journal of Computer Vision}, 131\penalty0 (2):\penalty0 531--551, 2023.

\bibitem[Shi et~al.(2021)Shi, Ye, Chen, Chen, Chen, and Kim]{monorcnn}
Xuepeng Shi, Qi Ye, Xiaozhi Chen, Chuangrong Chen, Zhixiang Chen, and Tae-Kyun Kim.
\newblock Geometry-based distance decomposition for monocular 3d object detection.
\newblock In \emph{ICCV}, pages 15172--15181, 2021.

\bibitem[Shi et~al.(2022)Shi, Guo, Mi, and Li]{stereo_centernet}
Yuguang Shi, Yu Guo, Zhenqiang Mi, and Xinjie Li.
\newblock Stereo centernet-based 3d object detection for autonomous driving.
\newblock \emph{Neurocomputing}, 471:\penalty0 219--229, 2022.

\bibitem[Simonelli et~al.(2019)Simonelli, Bulo, Porzi, L{\'o}pez-Antequera, and Kontschieder]{simonelli2019disentangling}
Andrea Simonelli, Samuel~Rota Bulo, Lorenzo Porzi, Manuel L{\'o}pez-Antequera, and Peter Kontschieder.
\newblock Disentangling monocular 3d object detection.
\newblock In \emph{ICCV}, pages 1991--1999, 2019.

\bibitem[Vaswani et~al.(2017)Vaswani, Shazeer, Parmar, Uszkoreit, Jones, Gomez, Kaiser, and Polosukhin]{attention}
Ashish Vaswani, Noam Shazeer, Niki Parmar, Jakob Uszkoreit, Llion Jones, Aidan~N Gomez, {\L}ukasz Kaiser, and Illia Polosukhin.
\newblock Attention is all you need.
\newblock \emph{NeurIPS}, 30, 2017.

\bibitem[Wang et~al.(2021)Wang, Du, Ye, Fu, Guo, Xue, Feng, and Zhang]{ddmp-3d}
Li Wang, Liang Du, Xiaoqing Ye, Yanwei Fu, Guodong Guo, Xiangyang Xue, Jianfeng Feng, and Li Zhang.
\newblock Depth-conditioned dynamic message propagation for monocular 3d object detection.
\newblock In \emph{CVPR}, pages 454--463, 2021.

\bibitem[Xu et~al.(2022)Xu, Zhong, and Neumann]{BtcDet}
Qiangeng Xu, Yiqi Zhong, and Ulrich Neumann.
\newblock Behind the curtain: Learning occluded shapes for 3d object detection.
\newblock In \emph{AAAI}, pages 2893--2901, 2022.

\bibitem[Yang et~al.(2022)Yang, Xu, Chen, Guo, Han, Ni, and Ding]{gpenet}
Fan Yang, Xinhao Xu, Hui Chen, Yuchen Guo, Jungong Han, Kai Ni, and Guiguang Ding.
\newblock Ground plane matters: Picking up ground plane prior in monocular 3d object detection.
\newblock \emph{arXiv preprint arXiv:2211.01556}, 2022.

\bibitem[Yin et~al.(2021)Yin, Zhou, and Krahenbuhl]{centerpoint}
Tianwei Yin, Xingyi Zhou, and Philipp Krahenbuhl.
\newblock Center-based 3d object detection and tracking.
\newblock In \emph{CVPR}, pages 11784--11793, 2021.

\bibitem[Yu and Koltun(2015)]{yu2015multi}
Fisher Yu and Vladlen Koltun.
\newblock Multi-scale context aggregation by dilated convolutions.
\newblock \emph{arXiv preprint arXiv:1511.07122}, 2015.

\bibitem[Yu et~al.(2018)Yu, Wang, Shelhamer, and Darrell]{DLA}
Fisher Yu, Dequan Wang, Evan Shelhamer, and Trevor Darrell.
\newblock Deep layer aggregation.
\newblock In \emph{CVPR}, pages 2403--2412, 2018.

\bibitem[Zhang et~al.(2023)Zhang, Qiu, Wang, Guo, Cui, Qiao, Li, and Gao]{monodetr}
Renrui Zhang, Han Qiu, Tai Wang, Ziyu Guo, Ziteng Cui, Yu Qiao, Hongsheng Li, and Peng Gao.
\newblock Monodetr: Depth-guided transformer for monocular 3d object detection.
\newblock In \emph{ICCV}, pages 9155--9166, 2023.

\bibitem[Zhang et~al.(2021)Zhang, Lu, and Zhou]{monoflex}
Yunpeng Zhang, Jiwen Lu, and Jie Zhou.
\newblock Objects are different: Flexible monocular 3d object detection.
\newblock In \emph{CVPR}, pages 3289--3298, 2021.

\bibitem[Zhang et~al.(2022)Zhang, Zheng, Zhu, Huang, Du, Zhou, and Lu]{dae}
Yunpeng Zhang, Wenzhao Zheng, Zheng Zhu, Guan Huang, Dalong Du, Jie Zhou, and Jiwen Lu.
\newblock Dimension embeddings for monocular 3d object detection.
\newblock In \emph{CVPR}, pages 1589--1598, 2022.

\bibitem[Zhou et~al.(2019)Zhou, Wang, and Kr{\"a}henb{\"u}hl]{centernet}
Xingyi Zhou, Dequan Wang, and Philipp Kr{\"a}henb{\"u}hl.
\newblock Objects as points.
\newblock \emph{arXiv preprint arXiv:1904.07850}, 2019.

\bibitem[Zhou et~al.(2021)Zhou, He, Zhu, Wang, Li, and Jiang]{zhou2021monoef}
Yunsong Zhou, Yuan He, Hongzi Zhu, Cheng Wang, Hongyang Li, and Qinhong Jiang.
\newblock Monoef: Extrinsic parameter free monocular 3d object detection.
\newblock \emph{IEEE Transactions on Pattern Analysis and Machine Intelligence}, 44\penalty0 (12):\penalty0 10114--10128, 2021.

\bibitem[Zhou et~al.(2023)Zhou, Zhu, Liu, Chang, and Guo]{monoatt}
Yunsong Zhou, Hongzi Zhu, Quan Liu, Shan Chang, and Minyi Guo.
\newblock Monoatt: Online monocular 3d object detection with adaptive token transformer.
\newblock In \emph{CVPR}, pages 17493--17503, 2023.

\bibitem[Zhu et~al.(2023)Zhu, Ge, Wang, and Peng]{monoedge}
Minghan Zhu, Lingting Ge, Panqu Wang, and Huei Peng.
\newblock Monoedge: Monocular 3d object detection using local perspectives.
\newblock In \emph{Proceedings of the IEEE/CVF Winter Conference on Applications of Computer Vision}, pages 643--652, 2023.

\end{thebibliography}
